\documentclass[10pt]{article}

\usepackage{amsmath, amsfonts, amssymb, mathrsfs, amsthm}
\usepackage{lineno}
\usepackage{hyperref}

\modulolinenumbers[5]

\usepackage{wrapfig}

\def\drawplusplus#1#2#3{\hbox to 0pt{\hbox to #1{\hfill\vrule height #3 depth
      0pt width #2\hfill\vrule height #3 depth 0pt width #2\hfill
      }}\vbox to #3{\vfill\hrule height #2 depth 0pt width
      #1 \vfill}}
\usepackage[all]{xy}
\usepackage{graphicx}

\newtheorem*{thm*}{Theorem}
\newtheorem*{mydef*}{Definition}
\newtheorem*{mylemma*}{Lemma}
\newtheorem*{myconjecture*}{Conjecture}

\begin{document}

%\begin{frontmatter}

\title{Paraconsistent Foundations\\
for Quantum Probability}

%% Group authors per affiliation:
\author{Ben Goertzel}

%\institute{OpenCog Foundation}
%\fntext[myfootnote]{Hanson Robotics}

%% or include affiliations in footnotes:
%\author[mymainaddress,mysecondaryaddress]{OpenCog Foundation}
%\ead[url]{ben@goertzel.org}

%\author[mysecondaryaddress]{Global Customer Service\corref{mycorrespondingauthor}}
%\cortext[mycorrespondingauthor]{Corresponding author}
%\ead{support@elsevier.com}

%\address[mymainaddress]{1600 John F Kennedy Boulevard, Philadelphia}
%\address[mysecondaryaddress]{360 Park Avenue South, New York}

\maketitle

\begin{abstract}
It is argued that a fuzzy version of 4-truth-valued paraconsistent logic (with truth values corresponding to True, False, Both and Neither) can be approximately isomorphically mapped into the complex-number algebra of quantum probabilities.  I.e., p-bits (paraconsistent bits) can be transformed into close approximations of qubits.   The approximation error can be made arbitrarily small, at least in a formal sense, and can be related to the degree of irreducible "evidential error" assumed to plague an observer's observations.   This logical correspondence manifests itself in  program space via an approximate mapping between probabilistic and quantum types in programming languages. 
\end{abstract}

%\end{frontmatter}

%\linenumbers

\tableofcontents

\section{Introduction}

The mathematics of quantum mechanics has been viewed and analyzed from a huge variety of different perspectives, each shedding light on different subtleties of its underlying structure and its connection to our everyday reality.   Here we add an additional thread to this conceptual polyphony, demonstrating a close connection between fuzzy paraconsistent logic and quantum probabilities.   This connection suggests  new variations on existing interpretations of quantum reality and measurement.  It also provides some tantalizing connections between the probabilistic and fuzzy logic used in modern AI systems and quantum probabilistic reasoning, which may have implications for quantum-computing implementations of logical inference based AI.

The ideas here arose as a spin-off from the work reported in \cite{goertzel2021paraconsistent}, which uses a variety of paraconsistent intuitionistic logic called Constructible Duality (CD) Logic as a means for giving a rigorous logic foundation to the PLN (Probabilistic Logic Networks) logic \cite{PLN} that has been used in the OpenCog AI project \cite{EGI1, EGI2} for well over a decade now.   Notation and concepts from  \cite{goertzel2021paraconsistent} are used liberally here, so the reader is basically required to ingest the relevant parts of that paper before this one.

Constructible Duality Logic features four-valued truth values which  in \cite{goertzel2021paraconsistent} are called p-bits, each of which may take any of the values: True, False, Both and Neither.   In \cite{goertzel2021paraconsistent}, an uncertain version of CD logic is created, involving tabulation and normalization of positive and negative evidence for a proposition in a novel way with elements of both probabilistic and fuzzy  reasoning.   Here one more step is taken, leveraging an extension of Knuth, Skilling and Goyal's \cite{goyal2010origin}  work on the foundations of quantum inference to {\it approximatively} map these uncertain CD truth values (p-bits) into complex numbers with a quantum-theory-friendly complex-probability interpretation.

Appropriately enough given the paraconsistent logic theme, our conclusions regarding the relation between uncertain paraconsistent logic and quantum probability arithmetic are both discouraging and exciting.   On the discouraging side, the approach reiterates the familiar reasons why there cannot ever be an exact isomorphism between fuzzy or probabilistic logic (in any of their current forms) and quantum logic or probability.  The crux of the matter (to simplify just a bit) is: Quantum algebra must be distributive, and fuzzy or probabilistic logic must use t-norm/conorm pairs for conjunction/disjunction -- but the only distributive t-norm/conorm pairs is min/max, which is not additively generative and thus can't be the basis of a morphism between paraconsistent logic operations and complex number operations.   

On the positive side, though, we do show that you can map between uncertain CD logic and quantum probability theory with small error -- most likely arbitrarily small error, though other strange things may possible arise as the error is shrunk all the way toward zero -- via a route that starts with assuming a certain small percentage of the observations underlying the truth values are erroneous.   There are t-norm/conorm pairs that closely approximate min/max (and are thus approximately distributive) and are also additively generative -- so using these to map fuzzy paraconsistent truth values into the complex plane, one leverages Knuth et al's work to obtain a result that fuzzy disjunction and conjunction approximately isomorphically map into complex number addition and multiplication.

Conceptually, this means one can interpret the use of complex arithmetic in quantum mechanics -- which is where most of the much-discussed "quantum weirdness" comes from -- as a different mathematical perspective on the use of p-bits to quantify observations.   If one admits that a proposition evaluated in a particular situation may sometimes be both True and False, or neither True nor False, rather than always being clearly on the True side or the False side -- and if one admits that every one of one's observations has a certain potential to be illusory or deceptive -- then one concludes that the algebra of one's truth values is approximately isomorphic to that of the complex plane.   

Whether the fuzzy paraconsistent logic view or the complex-probability view is the best way to look at a given situation then depends on what one needs to do.   Obviously for very many physics calculations, the complex-probability view is directly what one wants.   On the other hand if one is thinking about quantum AI or quantum biology, the story is less clear.   One of the tricky issues in quantum biology is the difficulty of drawing boundaries between quantum and classical portions of a system; the paraconsistent treatment of boundaries given in \cite{weber2010paraconsistent} and fuzzified/probabilized in \cite{goertzel2021paraconsistent} may become relevant.   

Regarding quantum computing for AI, the considerations here suggest it may be useful to look at mappings on the programming-language side that correspond to the mappings between logics given here.   Uncertain CD logic expressions map into pairs of types in dependent-type based programming languages which include probabilistic types.   It seems fairly clear how to create quantum types by analogy to these probabilistic types; and it also seems clear that, lifting the core ideas from this paper to the program world using the Curry-Howard correspondence, one obtains an approximate mapping from pairs of classical-probability-incorporating types to pairs of quantum-probability-incorporating types.   How this mapping cashes out in terms of practical quantum computing device, programming language or algorithm design is a wide open question.

\section{Fuzzy Paraconsistent Truth Values}

CD logic works with 4-valued truth values that we have in \cite{goertzel2021paraconsistent} called {\it p-bits} -- "paraconsistent bits", each of which has 4 possible values: True (1,0) , False (0,1), Both (1,1) or Neither (0,0).  

Given any logical language featuring constructs for True, False, $\land$, $\lor$, $\rightarrow$ and $\neg$, and a mapping from expressions in the language into a Heyting algebra $\mathcal{H}$, the Heyting algebra operations of meet, join and complement form an intuitionistic logic.   Patterson \cite{patterson1998implicit} shows that, similarly, if one has a mapping $h$ from expressions in the language into the product algebra $\mathcal{H} \times \mathcal{H}^{op}$ (where $\mathcal{H}^{op}$ denotes the opposite algebra to $\mathcal{H}$), then one obtains a CD logic with rules as follows.

\begin{itemize}
\item Basic mapping of logic operations
\begin{itemize}
\item $h(\alpha \land \beta) = h(\alpha) \sqcap h(\beta)$
\item $h(\alpha \lor \beta) = h(\alpha) \sqcup h(\beta)$
\item $h(\alpha \rightarrow \beta) = h(\alpha) \rightarrow h(\beta)$
\item $h(\alpha ) = \neg h(\alpha)$
\end{itemize}
\item Mapping of Four Units
\begin{itemize}
\item $h(\textrm{True}) = (1,0)$
\item $h(\textrm{False}) = (0,1)$
\item $h(\textrm{Neither}) = (0,0)$
\item $h(\textrm{Both}) = (1,1)$
\end{itemize}
\item  Logical operation across coordinates
\begin{itemize}
\item $(x, x')  \sqcap (y,y') = (x \sqcap y, x' \sqcup y')$
\item $(x, x')  \sqcup (y,y') = (x \sqcup y, x' \sqcap y')$
\item $(x, x')  \rightarrow (y,y') = (x \rightarrow y, x' \sqcap y')$
\item $\neg (x, x')  =  (x',x) $
\end{itemize}
\end{itemize}

CD logic is crisp, but we show in \cite{goertzel2021paraconsistent} that it can naturally be probabilized, via considering an ensemble of $N$ micro-situations, in each of which a certain proposition may be evaluated to have any of the four CD truth values.   One then associates with the proposition a 2D count value $(n^+, n^-)$ , where $n^+$ denotes the number of micro-situations in which there is positive evidence for the proposition (True or Both truth values), and $n^+$ denotes the number of micro-situations in which there is negative evidence for the proposition (False or Neither truth values).   The 2D count can be normalized into $(w^+, w^-) = (\frac{n^+}{N}, \frac{n^-}{N})$, yielding a paraconsistent uncertain truth value where the first component measures the amount of positive evidence and the second measures the amount of negative evidence, and the sum of the two components may vary from 0 to 2.

Given a t-norm/conorm pair $\mathcal{F} = (\top,\bot)$, one can define the operations $\sqcap_\mathcal{F}$ and $\sqcup_\mathcal{F}$ via

\begin{align*}
(w_1^+, w_1^-)  \sqcap_\mathcal{F} (w_2^+, w_2^-) = ( \top(w_1^+ , w_2^+) , \bot( w_1^- , w_2^-) ) \\
(w_1^+, w_1^-)  \sqcup_\mathcal{F} (w_2^+, w_2^-) = ( \bot(w_1^+ , w_2^+),  \top( w_1^-, w_2^-)  )\\
\neg_\mathcal{F} (w^+, w^-)  =  (w^-, w^+) 
\end{align*}

Note that here we carry over the negation operator from CD logic rather than using $\neg_\mathcal{F} (w^+, w^-)  =  (1-w^+, 1-w^-)$.   This means we don't have a nice algebraic relationship within each coordinate, but we get $( \sqcap_\mathcal{F} , \sqcup_\mathcal{F}, \neg_\mathcal{F})$ to be a DeMorgan triplet on 2D pairs, as $\sqcap_\mathcal{F}$ and $\sqcup_\mathcal{F} $ form a t-norm / conorm pair, $\neg_\mathcal{F} $ is an involutive negator, and 

$$
\forall w^+, w^- \in [0, 1]: \neg_\mathcal{F} \sqcup_\mathcal{F} (w^+, w^-) =\sqcap_\mathcal{F}( \neg_\mathcal{F}( w^+),  \neg_\mathcal{F}( w^-))
$$

\section{From Primitive Symmetries to Quantum Probabilities}  \label{sec:knuth}

Knuth, Skilling and Goyal \cite{goyal2010origin} present an elegant argument starting from some basic symmetries on combinations of observation sequences and ending up with quantum probabilities as the uniquely appropriate way to manage 2D truth values describing observation sets.  Conceptually and in some formal respects these arguments are a 2D extension of those in Knuth and Skilling's paper "Foundations of Inference" \cite{Knuth2012} which derives probability theory and information theory via basic symmetry arguments.

They consider an n-ary tree whose nodes above the leaf level represent composite objects, and where the objects represented by the children of a node are interpreted as a partition of the object represented by the node.   Looking at a function $p$ that measures the size of a node in the tree or a descending path in the tree, they explore the algebraic implications of some basic symmetry properties (commutativity, associativity, nontrivial dependency of a binary function on both variables, etc.) for the functional form of $p$.

First they argue that a few basic symmetry properties imply the rule

$$
p(B \bigoplus C) = p(B) + p(C)
$$

\noindent for disjoint destinations $B$ and $C$ from a binary source node $A = B \bigoplus C$, where e.g. $p(B)$ denotes the uncertainty quantification associated with $B$.   This is roughly the same as the argument from \cite{Knuth2012}  except that here the uncertainty quantifications are assumed 2D rather than 1D.

This part of their argument rests on a set of classical theorems arguing that any combinational operator one might use for $ \bigoplus$, if it satisfies a few reasonable-looking properties, must be isomorphic to the standard component-wise vector arithmetic operator +, in the sense that 

$$
p(B \bigoplus C) = f^{-1}( f( p(B) ) + f( p(C) ) )
$$

\noindent The argument is that if $\bigoplus$ is isomorphic to + in this sense, then we may as well consider it as actually being +, acting on a space of uncertainty values rescaled by $f$. \footnote{Philosophically, this is an interesting application of the Univalence Principle from homotopy type theory that "equals is equal to equivalence"; there is likely also a formal connection but we will not explore this here. }

Interestingly, not all conorm operators used to implement $\bigoplus$ (disjunction) operations in fuzzy logic actually obey the properties needed to make this sort of isomorphism work.   But many do, and these are typically called "additively generative" ones \cite{klement2013triangular}.

The next part of their argument moves on from addition to multiplication.   Where $\overrightarrow{UV}$ denotes a descending path from $U$ to $V$ in the tree, and $\circ$ denotes the concatenation of paths, they show that basic symmetry properties imply

\begin{eqnarray*}
p( \overrightarrow{UV} \circ \overrightarrow{VW} ) = p( \overrightarrow{UV})  * p( \overrightarrow{VW} )  \\
p(B) = p(\overrightarrow{BA}) p(\overrightarrow{AO}) p(O) \\
\end{eqnarray*}

\noindent where $O$ is the root object of the tree.

Key among these basic symmetry properties are left and right distributivity between $\circ$ and $\bigoplus$ -- which, notably, do not hold for any of the t-norm/t-conorm pairs used to quantify conjunction and disjunction in fuzzy logic, except for min/max (which are not additively generated).  So again we see that the "basic" symmetry properties used are actually quite restrictive in some relevant contexts.   This point gets at the heart of our unique contribution here, which is a way to partially dodge these issues via using a fuzzy conjunction / disjunction quantification that approximatively fulfills all the symmetries one wants, even if it fails to do so exactly.

Applying these arguments and further related ones to functions $p$ mapping paths into ordered pairs of reals, they arrive at the conclusion that the addition and multiplication involved in the above relationships must be the standard ones used in the complex number system.   From this point they proceed in a similar vein to Youssef \cite{Youssef1994} (though with different particulars), moving forward to Hilbert space and quantum mechanics from the assumption of complex-valued uncertainty quantifications. \footnote{Indeed Youseff's \cite{Youssef1994} analysis of quantum probabilities in terms of the Frobenius Theorem could be used as an alternate basis for the discussion here, with some technical differences but leading to the same basic point.}

A final point to make in reviewing the Knuth / Skilling / Goyal work is that explicit consideration of negation plays no role in their derivations -- they are concerned fundamentally with symmetries regarding conjunction and disjunction and their interrelationship.   This is an interesting contrast to intuitionistic logic where negation is the most subtle and vexed of the logical operations.   It is also convenient in the context of mapping paraconsistent logic into quantum probabilities, as we'll see below.

\section{An Approximate Mapping from Fuzzy Paraconsistent Truth Values to Quantum Probabilities}

What do we mean by an approximate mapping from fuzzy paraconsistent truth values into quantum probabilities?  Suppose $ (w_1^+, w_1^-)$ and $(w_2^+, w_2^-) $ represent bodies of evidence drawn from distinct sets of micro-situations; our aim here is to find a mapping function $\sigma$ defined by some t-norm/conorm pair $\mathcal{F} = (\top,\bot)$ with additive generating functions $(f,f^*)$ so that

$$
\sigma(a,b) = (f(a),f^*(b))
$$

\noindent and

\begin{align}
\sigma( (w_1^+, w_1^-) \sqcap_\mathcal{F} (w_2^+, w_2^-) ) \approx  \sigma (w_1^+, w_1^-) +  \sigma (w_2^+, w_2^-) \label{mc1} \\
\sigma( (w_1^+, w_1^-) \sqcup_\mathcal{F} (w_3^+, w_3^-) )  \approx  \sigma (w_1^+, w_1^-)  *  \sigma (w_3^+, w_3^-)\label{mc2} \\
\sigma( \neg_\mathcal{F}(w_1^+, w_1^-) )  \approx \neg \sigma(w_1^+, w_1^-) \label{mc3} 
\end{align}

\noindent where the operations on the lhs are defined as above and the operations on the rhs are standard complex number arithmetic, except for $\neg$ which is a simple reversal of coordinates ($\neg (a,b) = (b,a)$, or in complex number notation $\neg z = i z^*$ ).

It should be clear from the discussion in Section \ref{sec:knuth} why it's impossible to reduce the $\approx$ to a true equality.   The algebra on the right hand side is distributive, and the only t-norm/t-conorm pair that's distributive is max/min.   But for Equation \ref{mc1} to work as a precise equality, the operator $\sqcap_F$ needs to be additively generated, and max isn't. 

However, this argument doesn't prevent us from making the relationships hold with  $\approx$ and a small approximation error.   In \cite{goertzelprobability} we have presented an approximative version of Dupre' and Tipler's \cite{dupre2006cox} derivation of the rules of standard 1D probability theory from basic symmetries, showing that if one has operators that approximately obey the symmetries, then these operators must generally be the same as the standard probabilistic operators within a close degree of approximation.   Though the formal details will be different, it seems clear that a similar methodology will work with regard to Goyal, Skilling and Knuth's arguments going from basic symmetries to complex-arithmetic-based 2D probability theory rules.

To see how this sort of approximative argument can be leveraged here, we turn to the Schweizer-Sklar (SS) t-norms \cite{schweizer2011probabilistic}, defined via

$$
\top^{\mathrm{SS}}_p(x,y) = \begin{cases}
  \top_{\min}(x,y)          & \text{if } p = -\infty \\
  (x^p + y^p - 1)^{1/p}          & \text{if } -\infty < p < 0 \\
  \top_{\mathrm{prod}}(x,y)         & \text{if } p = 0 \\
  (\max(0, x^p + y^p - 1))^{1/p} & \text{if } 0 < p < +\infty \\
  \top_{\mathrm{D}}(x,y)            & \text{if } p = +\infty.
\end{cases}
$$

\noindent where

\begin{itemize}
\item $\top_{\mathrm{min}}(a, b) = \min \{a, b\}$ \\
\item $\top_{\mathrm{prod}}(a, b) = a \cdot b$  \\
\item $\top_{\mathrm{D}}(a, b) = \begin{cases}
  b & \mbox{if }a=1 \\
  a & \mbox{if }b=1 \\
  0 & \mbox{otherwise.} 
\end{cases}$
\end{itemize}

An additive generator for $\top^{\mathrm{SS}}_p$ for $\infty < p < \infty$ is

\begin{eqnarray*}
f^{\mathrm{SS}}_p (x) = \begin{cases} 
  -\log x           & \text{if } p = 0 \\
  \frac{1 - x^p}{p} & \text{otherwise.}
  \end{cases}
\end{eqnarray*}

Note that as $p \rightarrow -\infty$, $\top^{\mathrm{SS}}_p(x,y)  \rightarrow min(x,y)$.    Since for negative $p$ that are large in absolute value, $T^{\mathrm{SS}}_p(x,y)  \approx min(x,y)$, we also have that $\top^{\mathrm{SS}}_p(x,y)$ approximately obey the laws that $min(x,y)$ obeys exactly -- such as obeying distributive laws when considered together with its t-conorm.   For these $p$, therefore, the SS t-norm/t-conorm pair defines a mapping $\sigma$ so that

$$
\sigma(a,b) = (f^{\mathrm{SS}}_p(a), 1-f^{\mathrm{SS}}_p(1-b))
$$

\noindent which has properties

\begin{enumerate}
\item Equation \ref{mc1} holds precisely, since $\top^{\mathrm{SS}}_p$ is additively generated
\item Equation \ref{mc2} holds approximately, since $\top^{\mathrm{SS}}_p$ is approximately distributive
\item Equation \ref{mc3} holds precisely  
\end{enumerate}

So if we take $p$ that approaches $-\infty$ but doesn't get there yet, then we have a mapping function $\sigma$ that maps the CD logic of p-bits with SS t-norm/conorm {\it approximately isomorphically} into the complex arithmetic operators.

There is however an important subtlety regarding negation.   The CD negation operation maps into the reflection operator $z \rightarrow i z^*$, whereas the complex addition operator obviously works with the negation $z \rightarrow -z$.   If we map the latter back into the CD domain, it maps into the coordinatewise negation operator $\neg_c (w^+, w^-) = (1-w^+, 1-w^-)$ -- which does not properly form a DeMorgan triple with $\top^{\mathrm{SS}}_p$ , and doesn't agree with crisp CD negation in the case where the truth values are crisp.   

The mapping from conjunction / disjunction to complex number multiplication / addition doesn't require any considerations involving negation, so as far as this approximate isomorphism is concerned, one is free to consider negation however one wishes.    If one takes an intuitionistic-logic-like perspective, one can say that on both sides there are multiple relevant negation-type operators, each of which has different useful algebraic properties.

One more relevant detail to note is that Goyal, Skilling and Knuth \cite{goyal2010origin} initially find three different multiplication operators that seem almost acceptable as conjunctions for 2D probability values based on basic algebraic requirements like associativity, commutativity and so forth.  They then narrow down to the traditional complex-number multiplication based on some modestly subtle arguments regarding the results that the three candidates give in some example cases of multiple sequential and parallel measurements with natural symmetries (varying on standard Stern-Gerlach experiments).   

So summing up, the situation is that: The SS t-norm/co-norm approximates distributivity closely, and also supports additive generativity, thus allowing an isomorphic mapping that {\it approximately} takes paraconsistent disjunction into complex number addition, and  {\it approximately}  takes paraconsistent conjunction into what, after considering some additional technical criteria derived from physical sensibleness, is required to be complex number multiplication.

\section{Approximate Paraconsistent/Quantum Mapping via Evidential Error Uncertainty}

We've presented the approximate mapping from the paraconsistent world into the quantum world from a formal math perspective -- but what sense does it make conceptually?

There is a intriguing connection between the approximate mapping presented above and the semantics of uncertainty as regards existential and universal quantifiers leveraged in deriving PLN truth value formulas for these operators \cite{PLN} .   To model the uncertainty of these quantifiers in PLN, what has been done is to assume that every observation underlying every truth value estimate has a certain probability of being erroneous.   

So for instance, suppose every single one of one's 100 observations of rocks concur with the proposition that rocks are hard.   There are multiple sorts of uncertainty involved here.  One is that these 100 rocks might end up not to be representative of the overall population of rocks.  This kind of uncertainty is taken into account in the imprecise and indefinite probabilities used in the PLN framework, which involve confidence-weights attached to probability estimates, and formulas for assessing these weights based on evidence counts $n$.   There is also another kind of uncertainty though, which we may call "evidential error (EE) uncertainty"  -- there is the possibility that some of the observations of rocks being hard, were actually wrong ... that sometimes one perceived a soft rock to actually be hard, due to some sort of error in one's perceptual systems or a hack into the simulation running our physical reality, or whatever.   The probability of this kind of error pushes one to consider one's 100 observations of hard rocks as perhaps actually reflecting an observation of 100 rocks of which $n-k$ are hard and $k$ are not hard, where generally $\frac{k}{n}$ is small but not zero.

In PLN theory the consideration of EE uncertainty pushes one to model the uncertainties of universal and existential quantifiers using third-order probability distributions.   A similar mode of thinking seems to make sense in the current context.

If some of the observations underlying a p-bit $(w^+, w^-)$ are incorrect, then one views the pair $(n^+, n^-)$ as actually representing a distribution $\mathcal{P}_{(n^+, n^-)}$ over a set of values $(n^+ -k, n^- -k)$ where $k$ may be positive or negative and generally one needs $k 
\ll n = n^+ + n^-$ to have meaningful inferences.   Under the simplest assumptions the mean of the distribution $\mathcal{P}_{(n^+, n^-)}$ will be $\mu_{\mathcal{P}_{(n^+, n^-)}} =(n^+, n^-)$.

To cash out the implications of this expanded view of $(n^+, n^-)$  semantics for conjunction and disjunction operators, one could define e.g. a conjunction operator

$$
\mathcal{P}_{(n_1^+, n_1^-)} \sqcap_P \mathcal{P}_{(n_2^+, n_2^-)}
$$

\noindent which maps distributions to distributions.   One could then define an intersection operator on p-bits via projecting to and from these distributions,

$$
(n_1^+, n_1^-) \sqcap_P^* (n_2^+, n_2^-) = \mu_{ \mathcal{P}_{(n_1^+, n_1^-)}}  \sqcap_P \mu_{\mathcal{P}_{(n_2^+, n_2^-)}}
$$

\noindent (with some attention to make sure $\sqcap_P^*$ fulfills the t-norm requirements).   (While we have used $(n^+, n^-)$ language here, porting this to $(w^+, w^-)$ language is immediate.)

Exactly what this distribution will look like, will of course depend on how the distributions $\mathcal{P}_{(n^+, n^-)}$ are constructed.   However, qualitatively as  $\frac{k}{n} \rightarrow 0$, $\sqcap_P^*$ will behave similarly to $T^{\mathrm{SS}}_p$ as $p \rightarrow -\infty$.   I.e. roughly speaking $p' = \frac{n}{k}$ for $\sqcap_P^*$ will behave much like $p$ does for the SS t-norm.

From this view, the SS t-norm can be viewed as a heuristic approximation for $\sqcap_P^*$, valuable for initial exploration because of its simple analytical form.   The path from EE uncertainty to $\sqcap_P^*$ provides a conceptual framework in which the approximate mapping between paraconsistent and quantum logic naturally emerges.   

The conceptual story then looks like: {\bf Four-valued paraconsistent logic under small amounts of evidential-error uncertainty maps into quantum probability, within a close degree of approximation.}

\section{Toward Paraconsistent Programming with Quantum Type Theory}
\label{sec:CD-logic}

Warrell \cite{warrell2016probabilistic} has proposed a simple and appealing way of extending standard dependent type theory to include additional probabilistic types of interest for advanced AI.    The basic concept is to augment standard constructive type theory with a new primitive $\textrm{random}_\rho $ which denotes sampling from a Bernoulli distribution, and which comes along with some (obvious) rules for probabilistically-weighted beta reduction on pseudo-expressions in the dependent type language \footnote{These are referred to as pseudo-expressions because they are not necessarily type-correct; the reduction rule can be applied even if the "type pseudo-expression" $\tau$ is not type-correct}.

\begin{eqnarray*}
\textrm{random}_\rho ( \tau) \rightarrow^\rho_\beta ( \tau \textrm{ true}) \\
\textrm{random}_\rho ( \tau) \rightarrow^{1-\rho}_\beta ( \tau \textrm{ false}) 
\end{eqnarray*}

\noindent which has the meaning that the type of the expression $\textrm{random}_\rho ( \tau)$ inherits from $\tau$ with with probability $\rho$ and inherits from $~\tau$ with probability $1 - \rho$.

In \cite{goertzel2021paraconsistent} we have extended Warrell's approach via introducing constructs like

\begin{eqnarray*}
\textrm{random}_S ( \tau)  \\
\textrm{random}_M( \tau) 
\end{eqnarray*}

\noindent where $S$ represents a situation randomly chosen from an ensemble thereof, or $M$ represents a sub-metagraph drawn from a larger assumed metagraph. 

The mapping described here would allow one to approximatively replace e.g. $\textrm{random}_S ( \tau)$ with $\textrm{qrandom}_S ( \tau)$ which chooses $S$ from a (complex number) amplitude distribution rather than a conventional real-number probability distribution.   The approximate isomorphism explored here would lead to an approximate isomorphism between programs with probabilistic types corresponding to paraconsistent logic expressions, and programs with quantum types corresponding to quantum logical expressions.

One may speculate that this provides a potentially interesting novel source of quantum algorithms -- by taking classical algorithms and "quantizing" them using EE-based or similarly-behaving transformations.  

\section{Future Directions}

The argument given here, while conceptually compelling and apparently mathematically sound, is also somewhat sketchy as presented.   One suspects that filling in all the details in a fully rigorous way will lead to some additional discoveries as well as perhaps more clarity on the limitations of the approach.

The core message that complex-valued quantum probability can be approximatively viewed as a transformed version of 4-valued paraconsistent logic -- that qubits can be viewed essentially as noisy transformed p-bits -- is one that seems to have some subtlety to it from a philosophical and formal-logic view.   Whether this approach to the quantum world has any concrete practical value remains to be seen and there is clearly a fairly long road toward mining any such value.   We have the intuition that quantum biology, quantum computing and especially quantum AI may be domains in which such exploration could be fruitful.

\bibliographystyle{alpha}
\bibliography{bbm}

\end{document}